\documentclass[letterpaper, 10 pt, journal, twoside]{ieeetran}

\usepackage{graphics} 
\usepackage{graphicx}
\usepackage{bm}
\usepackage{amsmath,amsfonts}
\usepackage{amssymb}
\usepackage{multirow}
\usepackage{threeparttable}
\usepackage{array}
\newcolumntype{P}[1]{>{\centering\arraybackslash}p{#1}}
\usepackage{tikz}
\newcommand*\circled[1]{\tikz[baseline=(char.base)]{
            \node[shape=circle,draw,inner sep=0.01pt] (char) {#1};}}
\usepackage{pdfpages}
\usepackage{lettrine}
\usepackage{dblfloatfix}
\usepackage{diffcoeff,amssymb}
\usepackage{graphicx}
\usepackage[export]{adjustbox}
\usepackage{hhline}
\usepackage{tabularray}
\usepackage{cite}
\usepackage{wasysym}
\usepackage[switch]{lineno}

\ifCLASSINFOpdf

\else

\fi

\title{\LARGE 
The JSI-KneExo: Active, Quasi-Passive, Pneumatic, Portable Knee Exo with Bidirectional Energy Flow for Air Recovery in Sit-Stand Tasks}

\author{Luka Mišković$^{1}$ Tilen Brecelj$^{1}$  Miha Dežman$^{2}$ Tadej Petrič$^{1}$
\thanks{This work was supported by grants from the Slovenian
Research Agency; PR-10489, and N2-0153. Video attachment available along with the paper. } 

\thanks{$^{1}$Luka Mišković, Tilen Brecelj and Tadej Petrič are with the Department of Automatics, Biocybernetics and Robotics, Jožef Stefan
Institute, Jamova cesta 39, 1000 Ljubljana, Slovenia,
        {(Luka Mišković \tt\small luka.miskovic@ijs.si)}}%

\thanks{$^{2}$Miha Dežman is with the High Performance Humanoid Technologies (H2T), Institute for Anthropomatics and Robotics (IAR), Karlsruhe Institute of Technology (KIT),
Adenauerring 2, 76131 Karlsruhe, Germany
}
}

\begin{document}

\maketitle
\thispagestyle{plain}
\pagestyle{plain}

\begin{abstract}
While existing literature encompasses exoskeleton-assisted sit-stand tasks, the integration of energy recovery mechanisms remains unexplored. To push these boundaries further, this study introduces a portable pneumatic knee exoskeleton that operates in both quasi-passive and active modes, where active mode is utilized for aiding in standing up (power generation), thus the energy flows from the exoskeleton to the user, and quasi-passive mode for aiding in sitting down (power absorption), where the device absorbs and can store energy in the form of compressed air, leading to energy savings in active mode. The absorbed energy can be stored and later reused without compromising exoskeleton transparency in the meantime.
In active mode, a small air pump inflates the pneumatic artificial muscle (PAM), which stores the compressed air, that can then be released into a pneumatic cylinder to generate torque. All electronic and pneumatic components are integrated into the system, and the exoskeleton weighs 3.9 kg with a maximum torque of 20 Nm at the knee joint. The paper describes the mechatronic design, mathematical model and includes a pilot study with an able-bodied subject performing sit-to-stand tasks. The results show that the exoskeleton can recover energy while assisting the subject and reducing muscle activity.
Furthermore, results underscore air regeneration's impact on energy-saving in portable pneumatic exoskeletons.
\end{abstract}

\begin{figure*}[b!]
        \centering 
    \includegraphics[clip,trim=0cm 0cm 0cm 0cm,width=1\textwidth]{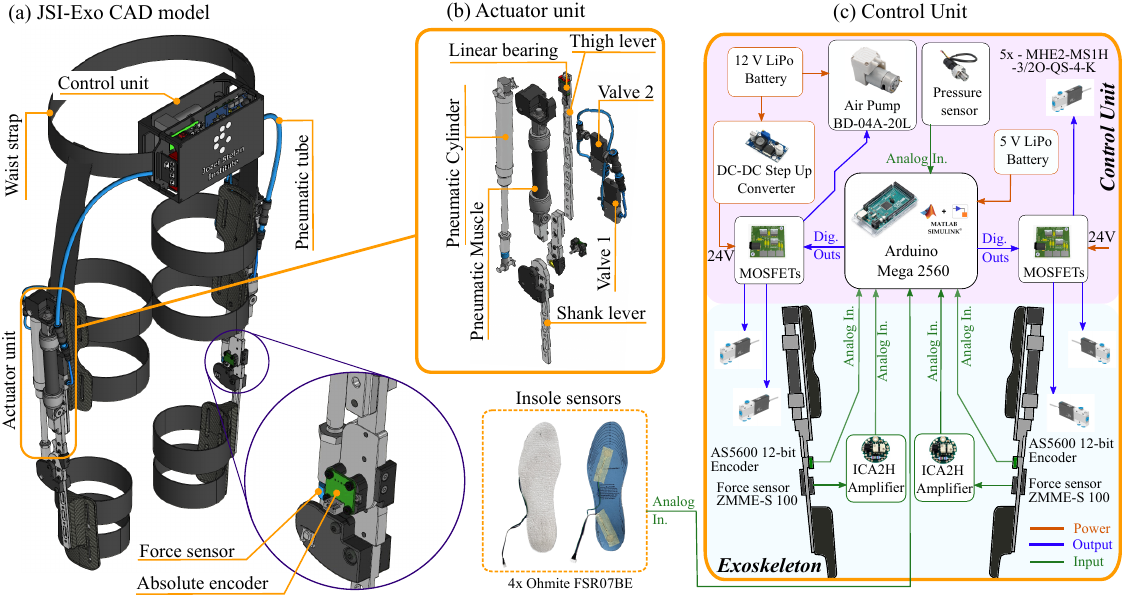}
    \caption{(a) The JSI-KneExo comprises three modular parts: control unit, left-leg exoskeleton, and right-leg exoskeleton. (b) The actuator unit incorporates air solenoid valves, PAM, pneumatic cylinder, force sensor, absolute encoder, shank and thigh lever, and a linear bearing facilitating linear PAM deflection. (c) Pink background highlights control unit components, while light blue denotes exoskeleton (leg side) components; optional insole sensors can be added.}\label{fig:overview}
\end{figure*}

Efficient energy utilization is a crucial consideration in every technical system. Human joints can both generate and absorb power \cite{DonaldA.Neumann2016KinesiologyRehabilitation}, which opens up the space for energy recovery. 
For example, regenerative exoskeletons are able to capture otherwise wasted human energy and produce electricity when backdriven by joint movements during negative power intervals \cite{Donelan2008BiomechanicalEffort, Wu2022GeneratingExoskeleton, Laschowski2019Lower-LimbReview, Dollar2008DesignRunning}.  
Another type of energy storing and reusing in exoskeletons resembles the one in the musculoskeletal system, and it is found in quasi-passive exoskeletons which absorb energy within elastic components during negative power intervals and subsequently supply torque to the joints, all without generating net positive work \cite{Collins2015ReducingExoskeleton, DiNatali2020PneumaticExoskeleton, Miskovic2022PneumaticApplications, Kumar2020ExtremumExoskeleton, Sutrisno2019EnhancingExoskeletons, Walsh2007AAugmentation}. Unlike fully passive designs, these exoskeletons contain clutches, which enable transparency when assistance isn't required.
While most of these devices primarily target energy recovery during walking, Laschowski et al. \cite{Laschowski2021SimulationRegeneration} highlighted the potential of energy recovery during sitting transfers where the hip, knee, and ankle power is almost entirely negative, while standing up requires positive joint power \cite{DonaldA.Neumann2016KinesiologyRehabilitation}. 
Existing literature covers exoskeletons-assisted sit-stand tasks, yet without any type of energy recovery \cite{Schmidt2017TheTransfers, Shepherd2017DesignAssistance, Liu2021LowSit-to-Stand, Zhu2021DesignOrthoses, Vantilt2019Model-basedMovements}.
An exception is found in \cite{Jamsek2020GaussianExoskeletons}, where a task with a quasi-passive exoskeleton involved sitting down while accumulating energy, and subsequently utilizing it for lifting. Despite the effort, the study revealed limitations, specifically the challenge of disengaging the clutch without compromising assistance for lifting and the inability to disengage the clutch under load.
Generally, in quasi-passive exoskeletons based on mechanical springs, achieving transparency after the energy has been accumulated necessitates mechanisms to detach the elastic element from the output lever, otherwise, the accumulated energy is lost. 
Even if achieving transparency is mechanically accomplished, the stored energy within mechanical elastic elements, like springs, cannot be stashed in an energy container for later use. 
However, this is easily achievable with pneumatic elastic components (air springs) as air serves as a lightweight elastic medium that can be passed to another energy reservoir and used later. 

In an optimal scenario, exoskeleton-assisted sit-stand task include: assistance with energy absorption (in the Exo) when sitting down, efficient energy storing for future use, enabling transparency without energy loss, and actively assisting in standing up using stored energy, which would reduce overall consumption. We hypothesize this is achievable using pneumatic technology, thus we design a portable pneumatic exoskeleton that operates as active, quasi-passive, and transparent while keeping energy. 
In the literature, pneumatic actuators have been utilized in the active \cite{Maeda2012MuscleMuscles, Knaepen2014Human-robotExoskeleton} and quasi-passive exoskeletons \cite{DiNatali2020PneumaticExoskeleton, Miskovic2022PneumaticApplications, Miskovic2023PneumaticEvaluation}. They are appealing due to their high weight-to-power ratio and inherent compliance, still, when used as active the necessity for an air tank poses portability challenges for pneumatic exoskeletons. Notably, a few prior studies have successfully devised active portable pneumatic exoskeletons integrating onboard air compressors and metal air tanks; \cite{Wehner2013AAssistanceb} weighing 9.12 kg, \cite{Heo2020BackdrivableAssistance} weighing 9.95 kg.
Our previous work developed a method for modulating air spring stiffness by drawing air from the atmosphere without an air pump or supply. The method was applied in the variable stiffness joint mechanism \cite{Miskovic2023PneumaticEvaluation}. In this paper, the JSI-KneExo is introduced with key \textbf{contributions} as follows:
\begin{enumerate}
\item Introducing a solution that enables bidirectional energy flow between the user and exoskeleton, allowing energy recovery and storage in the form of compressed air while simultaneously assisting the user. Energy recovery ultimately leads to overall energy savings. 
\item Design is enabled by an air regenerative actuator integrated into a lightweight, slim, and portable pneumatic knee exoskeleton that can switch between different modes, such as active, quasi-passive, or transparent which is mathematically described. 
\end{enumerate}

\section{THEORETICAL ANALYSIS} \label{Sec.:theor}


\subsection{Exoskeleton Design}
The JSI-KneExo (Fig. \ref{fig:overview}a) is a modular system made up of three major components: the left-leg exoskeleton with an actuator, the right-leg exoskeleton with an actuator, and the control unit. Depending on the application, one or both legs of the exoskeleton may be used simultaneously. The whole system weighs 3.9 kg (each leg of the exoskeleton 1.25kg and the control unit 1.4kg). The range of motion is 0 - 135$^{\circ}$.

\subsubsection{The actuator unit}

The pneumatic actuator (Fig.~\ref{fig:overview}b) comprises a pneumatic cylinder (DNSU-25-100-PPV-A, FESTO, Germany), which generates force when pressurized air enters its chamber. As the cylinder is attached to the shank and thigh levers, the force in the cylinder causes torque in the joint with a maximum value of 20 Nm at a maximum pressure of 8 bar. The cylinder has a maximum stroke of 100 mm and a bore diameter of 25 mm. The compressed air generated by the air pump is stored in the pneumatic muscle DMSP-20-100N-RM-RM (initial diameter 20mm, initial length 100mm), which when inflated shrinks its length and increases its diameter. The PAM's lower end is fixed, upper-end slides on the linear bearing. Each unit contains two solenoid air valves (MHE2-MS1H-3/2O-QS-4-K, FESTO, Germany) for state change, as detailed further in the paper. 
The absolute encoder AS5600 (ams-OSRAM AG, Austria) measures the joint angle, while the cylinder force is measured by the force sensor ZMME-S100 (Zhiminsensor, China). 

\subsubsection{The control unit}

The control unit is shown in (Fig. \ref{fig:overview}c). The JSI-KneExo is fully portable as the compressed air is supplied by the air pump BD-04A-20L, BodenFlo, China, driven by a 12 V LiPo battery. If running continuously, the pump has an autonomy of $\approx$3.4 h. Six \mbox{MOSFET} switches, arranged in two PCBs control solenoid valve positions. An extra solenoid valve is included in the control unit box, the function of which is explained in the next section. A pressure sensor is integrated to continuously monitor the pressure in both PAMs. The voltage is converted to 24 V for the solenoid valves by the DC-DC converter. The main microcontroller is an Arduino Mega 2560.

\subsection{Portable pneumatic design}

\begin{figure}[ht!]
        \centering 
    \includegraphics[clip,trim=0cm 0cm 0cm 0cm,width=0.45\textwidth]{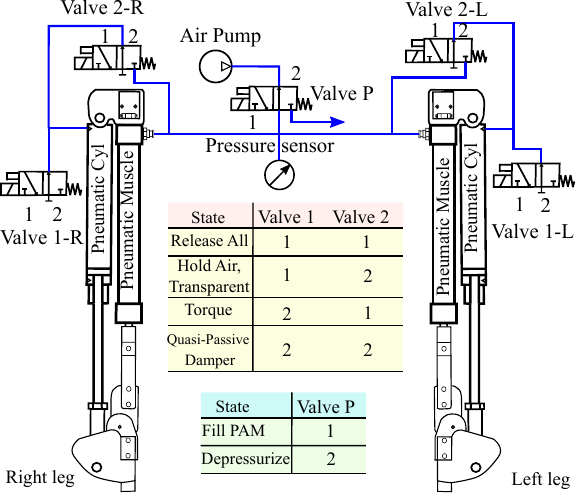}
     \caption{Pneumatic circuit of the JSI-KneExo. 
     Different valve positions achieve different exoskeleton states.
    }\label{fig:pneumatics}
\end{figure}

The portable pneumatic circuit is shown in Fig. \ref{fig:pneumatics}. All five solenoid valves are identical and monostable, transitioning smoothly between '1' and '2' positions with a $\approx $2 ms frequency. The air pump generates compressed air, which charges the PAM, and pressure is tracked using a sensor. 
The valve P depressurizes the tube after air pumping; otherwise, by restarting the pump the reverse pressure holds the pump back when it needs the most power. Symmetrical pneumatic connections exist on both sides (left and right). Different valve combinations achieve various states.
For example, in "Release all" both valves 1 and 2 are set to '1', allowing atmospheric pressure to enter the cylinder and PAM, resulting in air release. The second state, "Hold Air, Transparent," involves the air pump filling the PAM to the desired pressure set in the control algorithm, but valve 1 has a clutch function, meaning that if in position '1' the pneumatic cylinder appears transparent as the chamber is connected to the atmosphere and air cannot be compressed. After being in the transparent state, by setting valve 2 to '1', compressed air enters the cylinder chamber, producing force and joint torque. It's crucial to revert valve 1 to '2' to prevent air discharge. This state is called "Torque" or active mode.
The last state is the "Quasi-passive Damper" or quasi-passive mode, in which both valves are set to position '2'. Torque is generated as the air inside the cylinder is compressed by knee flexion, acting as a nonlinear progressive compression spring resisting rotation. Furthermore, in this state, compressed air can be returned to the PAM by flexing the joint and switching valve 2 to '1' so that air flows back into the PAM.

\subsection{Mathematical model}

\subsubsection{Air pump}

The air pump has the function of charging the PAM with compressed air to the desired pressure. Commercially available air pumps are rated for no-load air flow at 0 bar pressure. Therefore, it is hard to calculate the time required to fill the PAM as the pressure increases. Another approach involves experimentally identifying the pressure-time profile. 
\begin{figure}[ht!]
        \centering 
    \includegraphics[clip,trim=0cm 0cm 0cm 0cm,width=0.47\textwidth]{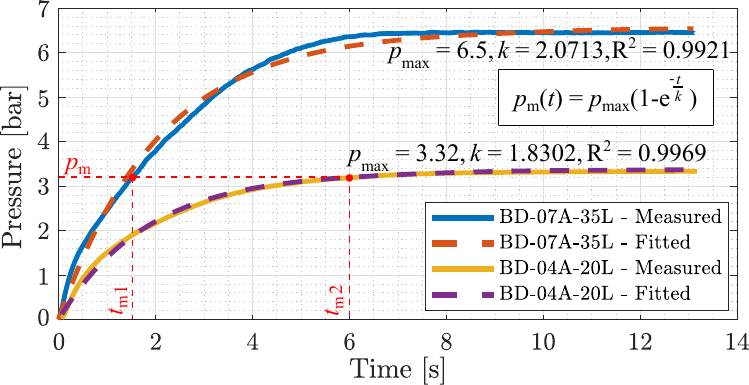}
    \caption{Filling compressed air into the PAM with two different pumps, the larger BD-07A-35L and the smaller BD-04A-20L. The equation fitted through measured data has the form: ${p_\text{m}(t)=p_\text{max}(1-e^{-t/k})}$. For the larger pump: $p_\text{max}=6.5$ bar, $k=2.0713$, and $\text{R}^2=0.9921$. For the smaller pump: \mbox{$p_\text{max}=3.32$ bar}, $k=1.8302$ and $\text{R}^2=0.9969$. }\label{fig:pump}
\end{figure}
Hence, two air pumps were experimentally identified and compared: BD-04A-20L (maximum pressure: 3.32 bar, 20 L/min flow rate at 0 bar) and the air piston pump BD-07A-35L (maximum pressure: 6.5 bar, 35 L/min flow rate at 0 bar).
Fig. \ref{fig:pump} depicts the measured pressure-time profile when filling the pneumatic muscle DMSP-20-100N-RM-RM for both pumps. The measured data were fitted with the following exponential function:

\begin{equation}\label{eq:pm}
{p_\text{m}(t)=p_\text{max}(1-e^{-\frac{t}{k}})}\,,
\end{equation}
where $p_\text{m}(t)$ is the actual pressure at time $t$, $p_\text{max}$ is the max. nominal pressure for the pump, and $k$ is the pump constant identified for both pumps. The coefficient of determination is R$^2=0.9921$ for BD-07A-35L and R$^2=0.9969$ for BD-04A-20L. 
The time $t_\text{m}$ required to reach the desired pressure $p_\text{m}$ in the PAM can now be determined using Eq. (\ref{eq:pm}).




\subsubsection{Pneumatic Artificial Muscle (PAM)}
When the air gets compressed inside the PAM, its length $L_\text{m}$ and volume $V_\text{m}$ change. The analytical dependency $V_\text{m}$($L_\text{m}$) is known and adopted from \cite{Martens2017ModelingModels}. 
However, with a known PAM pressure $p_\text{m}$ its length $L_\text{m}$ can only be estimated from the FESTO datasheet for the exact PAM or other existing PAM models \cite{Martens2017ModelingModels,Chou1996MeasurementMuscles}. 
Since FESTO does not provide an analytical model, the PAM model was determined experimentally.
The dependency between the pressure $p_\text{m}$, length $L_\text{m}$, and the volume $V_\text{m}$ is needed to compute the actuator's torque. Therefore, the model was determined experimentally by compressing air into the PAM and measuring the contraction $\varepsilon$. The PAM force was zero during the identification as the linear bearing allows unrestrained contraction.




\begin{figure}[ht!]
        \centering 
    \includegraphics[clip,trim=0cm 0cm 0cm 0cm,width=0.48\textwidth]{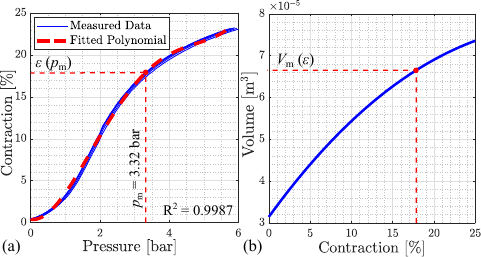}
    \caption{Model-identification of the PAM DMSP-20-100N-RM-RM. In (a) PAM contraction versus pressure. In (b) PAM volume versus contraction. }\label{fig:PAM}
\end{figure}
The experimental identification is shown in Fig. \ref{fig:PAM}, where (a) shows the measured contraction $\varepsilon$ as a function of PAM pressure $p_\text{m}$, and (b) shows the analytical model linking PAM volume $V_\text{m}$ and contraction $\varepsilon$, given in Eq. \ref{eq:Vm}, for different contractions where the maximum contraction for the chosen PAM is equal to 25$\%$.
The measured data were fitted with a fourth-degree polynomial, formulated as \begin{equation}\label{eq:epsilon}
    \varepsilon(p_\text{m})=c_{1}p_\text{m}^{4}+c_{2}p_\text{m}^{3}+c_{3}p_\text{m}^{2}+c_{4}p_\text{m}+c_{5}\,,
\end{equation}
where the coefficients are $c_{1}=0.1022, c_{2}=-1.3370, c_{3}=5.1426, c_{4}=-0.8131, c_{5}=0.4189$, with the coefficient of determination $\text{R}^2=0.9987$.
Using the identified relation $\varepsilon(p_\text{m})$, the contraction $\varepsilon$ and hence the length $L_\text{m}$ of the PAM for each pressure $p_\text{m}$ can be obtained as ${L_\text{m}}={L_\text{0}}-\varepsilon$, where ${L_\text{0}}$ is the initial PAM length. This also allows the determination of the PAM volume, $V_\text{m}$, for each length $L_\text{m}$, which was previously unknown.  
The PAM volume can be calculated according to \cite{Martens2017ModelingModels} as follows
\begin{equation}\label{eq:Vm}
{V_\text{m}}=\frac{{L_\text{m}}L_\text{f}^2-{L_\text{m}}^3}{4 \pi n_\text{tu}^2}\,,
\end{equation}
where $n_\text{tu}$ is the number of thread turns, and $L_\text{f}$ is the thread length calculated according to \cite{Chou1996MeasurementMuscles}. 
Now this relationship can be used to calculate the actuator parameters explained in the next subsection. 

\begin{figure}[ht!]
        \centering 
    \includegraphics[clip,trim=0cm 0cm 0cm 0cm,width=0.48\textwidth]{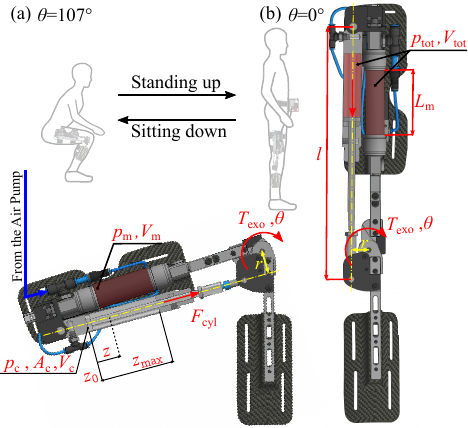}
    \caption{(a) Exoskeleton flexed at $\theta=107^\circ$, the angle where $z=z_0$ for $\varepsilon(p_\text{m}=3.32,\text{bar})$ contraction. (b) Exoskeleton at $\theta=0^\circ$ for user standing.}\label{fig:exo_model}
\end{figure}

\subsubsection{Actuator Pressure, Force, and Torque}

the mathematical model of the actuator is given for one side exoskeleton leg and is further derived according to Fig. \ref{fig:exo_model}, where the transition from (a)$\to$(b) implies standing up and reversely sitting down, meaning that in (a)$\to$(b) the pressure is released from the PAM and shared with the pneumatic cylinder, while in (b)$\to$(a) the pressure is returned to the PAM. The initial angle in (a) is $\theta=107^\circ$, as this is the angle at which the piston rod comes to the end of the cylinder, for contraction $\varepsilon(p_\text{m}=3.32\,\text{bar})$, which is identified as a maximum pressure for the selected air pump BD-04A-20L.  
At the point where the angle $\theta$ starts decreasing, the air valve connecting the cylinder and the PAM is switched to position '1' as explained earlier. This results in two volumes merging, and the common volume is defined as:
\begin{equation}
    V_\text{tot}=V_\text{c}+V_\text{m}\,,
\end{equation}
where $V_\text{c}$ is the volume of the cylinder with the volume of the pneumatic tube added and is equal to
\begin{equation}
    V_\text{c}=A_\text{c}z+A_\text{tc}l_\text{tc}\,, \quad z\in[z_0, z_{max}]\,,
\end{equation}
where $A_\text{tc}$ is the cross-sectional surface area of the pneumatic tube and $l_\text{tc}$ is its length from the cylinder to the PAM. The current position of the piston $z$ is expressed as
\begin{equation}
z=z_\text{max}\frac{l-l_\text{min}}{l_\text{max}-l_\text{min}}\,,
\end{equation}
where $l$ is the cylinder length (Fig. \ref{fig:exo_model}b), which was previously derived in our article \cite{Miskovic2023PneumaticEvaluation}, and $l_\text{min}$ is the length when the piston touches the end of the cylinder ($z=z_\text{0}$). The maximum actuator length $l_\text{max}$ is the length for $z=z_\text{max}$. 

The PAM volume was given in Eq. \ref{eq:Vm}, but now the volume of the pneumatic tube ($A_\text{tm}l_\text{tm}$) is added for the actuator as follows
\begin{equation}
{V_\text{m}}=\frac{{L_\text{m}}L_\text{f}^2-{L_\text{m}}^3}{4 \pi n_\text{tu}^2}+A_\text{tm}l_\text{tm}\,.
\end{equation}
Once the total volume is known, and isothermal expansion is assumed, the air pressure ${p_\text{tot}}$ expanding from the initial pressure $p_\text{tot}^{_\text{init}}$ and the initial volume $V_\text{tot}^{_\text{init}}$ to the current volume $V_\text{tot}$ can be computed as
\begin{equation}\label{eq:ptot}
{p_\text{tot}}={p_\text{tot}^{_\text{init}}}\frac{V_\text{tot}^{_\text{init}}}{V_\text{tot}}\,,
\end{equation}
As volume $V_\text{tot}$ gradually increases, pressure $p_\text{tot}$ decreases, which means that as the angle $\theta$ approaches $0^\circ$ (knee fully extended), $L_\text{m}$ also decreases, consequently affecting the cylinder length $l$, the current position of the piston $z$, and finally the volume of the cylinder $V_\text{c}$. Therefore, the PAM length must be updated for each new pressure during the transition according to the model identified in Eq. \ref{eq:epsilon}.

Once the current pressure is known during the transition, the cylinder force $F_\text{cyl}$ can be expressed as the pressure $p_\text{tot}$ acting on the piston surface area $A_\text{c}$, as follows
\begin{equation}\label{eq:force}
    F_\text{cyl}=p_\text{tot}A_\text{c}\,.
\end{equation}

\begin{figure}[ht!]
        \centering 
    \includegraphics[clip,trim=0cm 0cm 0cm 0cm,width=0.4\textwidth]{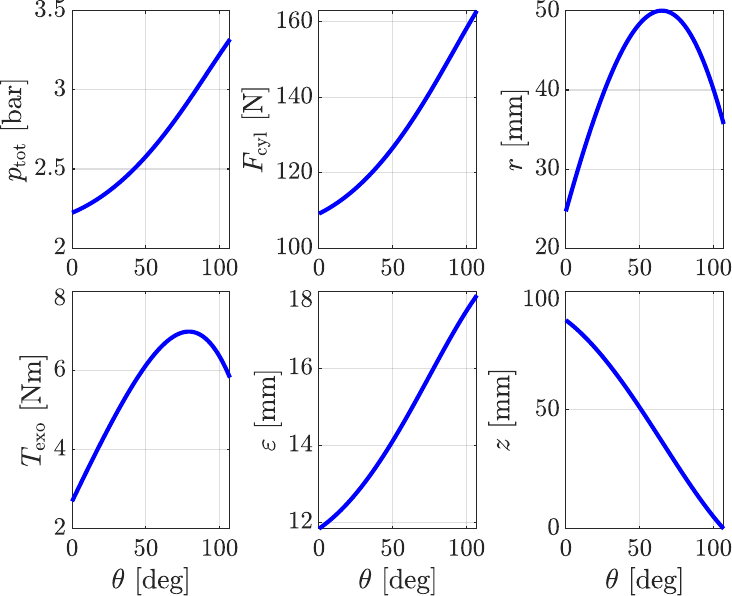}
    \caption{Theoretical profiles of different physical parameters for the joint rotation from $\mbox{$\theta=107^\circ$}$ to $\theta=0^\circ$. }\label{fig:theory}
\end{figure}

Finally, with the known cylinder force $F_\text{cyl}$, the exoskeleton torque $T_\text{exo}$ can be computed as follows
\begin{equation}
    T_\text{exo}=F_\text{cyl}r\,,
\end{equation}
where the lever arm $r$ (see Fig. \ref{fig:exo_model}) was mathematically modeled in our previous article \cite{Miskovic2023PneumaticEvaluation}. 
The computed values of different physical parameters for the joint rotation from $\theta=107^\circ$ to $\theta=0^\circ$ are shown in Fig.~\ref{fig:theory}.
In the case when the angle theta is changed from $\theta=0^\circ$ to $\mbox{$\theta=107^\circ$}$ (sitting down), the process is reversed if the "Torque" state is still active. Therefore, by rotating the joint back to $\mbox{$\theta=107^\circ$}$ the air is returned into the PAM while generating the same torque. 
However, when the "Quasi-passive Damper" state is activated (see Fig. \ref{fig:pneumatics}) at the angle of $\theta=0^\circ$ and the joint is rotated back to $\mbox{$\theta=107^\circ$}$, the air is compressed only in the cylinder's volume, thus the pressure and torque reach higher values than in the active mode. Still, the air can be returned back into the PAM by setting valve 2 in position '1' once the air has been compressed in the cylinder (person sitting). The mathematical model in this paper doesn't account for PAM stretching that occurs beyond $\theta=107^\circ$, even though rotation can reach $\theta=135^\circ$. Experimental results will assess the exoskeleton's overall dynamics.

 \section{The JSI-KneExo EVALUATION} \label{Sec.:experiment}

\begin{figure}[ht!]
        \centering 
    \includegraphics[clip,trim=0cm 0cm 0cm 0cm,width=0.4\textwidth]{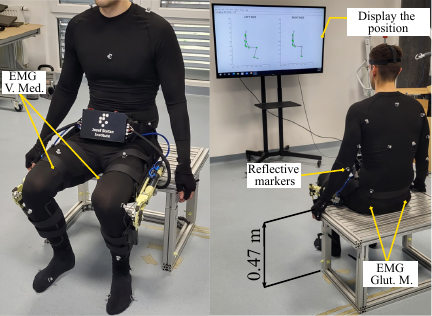}
    \caption{Experimental setup: EMG electrodes, reflective markers, and screen for visual feedback.}\label{fig:experimental_setup}
\end{figure}

This section presents a pilot study with an able-bodied subject (27 years old, 90 kg, 1.93 m tall), aiming to validate JSI-KneExo's key contributions. The exoskeleton was controlled in Simulink Real-Time™, while the pressure, angle, and force data were collected and analyzed offline in Matlab. 
The experimental setup is in Fig. \ref{fig:experimental_setup}. To determine exoskeleton impact on the subject, a multi-channel Trigno wireless EMG system (Delsys, USA) recorded Vastus Medialis (V. Med.) and Gluteus Maximum (Glut. M.) muscle activity at 2148.15 Hz, following SENIAM guidelines \cite{Hermens2000DevelopmentProcedures}. Prior to experiments, muscle maximum voluntary contractions (MVC) were recorded.
Optitrack's 16 cameras (NaturalPoint, USA) at 120 Hz captured joint kinematics. Using 37 markers as per Motive Optical motion capture software, the human skeleton was reconstructed. Visual feedback on a screen ensured subject positioning consistency for repeatable experiments.

\subsection{Experiment}
The subject was instructed to perform ten sit-to-stand and stand-to-sit transitions under unassisted (transparent) and assisted conditions. The experiment aims to validate the energy recovery (compressed air return to PAM) while relieving the leg muscles, and to evaluate the exoskeleton operation, highlighting the transparency achieved after every sit-down while storing recovered air. When standing up, the exoskeleton assisted the subject in active mode, and when sitting down, the exoskeleton assisted the subject in quasi-passive mode. 
In the unassisted trial, the state was set to "Release All" to enable entire transparency.

\begin{figure}[ht!]
        \centering 
    \includegraphics[clip,trim=0cm 0cm 0cm 0cm,width=0.48\textwidth]{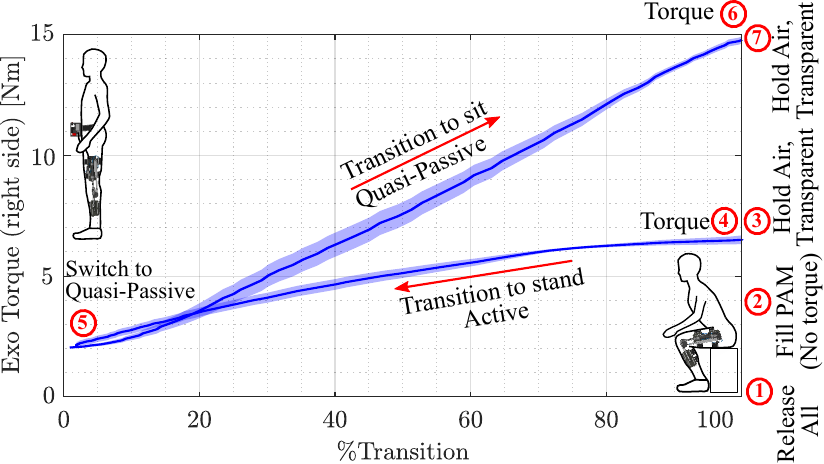}
    \caption{Right leg exoskeleton torque during transition, scaled from 100\% (sitting) to 0\% (standing), with numbered circles indicating different states.}\label{fig:torque_vs_trans}
\end{figure}

\begin{figure}[ht!]
        \centering 
    \includegraphics[clip,trim=0cm 0cm 0cm 0cm,width=0.48\textwidth]{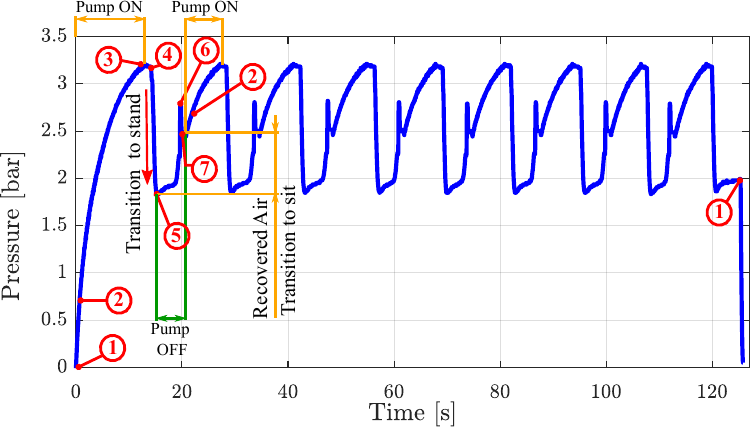}
    \caption{Pressure sensor measurements over time with state distinctions using numbers. Pump on-off intervals and air recovery periods are marked.}\label{fig:pressure_vs_time}
\end{figure}

{\underline{Strategy and Control:}}
The controller for the assisted condition involves a state machine with states triggered based on threshold values from a pressure sensor and two encoders, with knee angular velocity derived through knee angle numerical differentiation.
Exoskeleton torque was computed offline by multiplying the cylinder force with the lever arm length, calculated from the knee angle and mathematical model. 
In Fig. \ref{fig:torque_vs_trans}, the average exoskeleton torque (right leg) and its standard deviation across ten repetitions are presented in relation to the transition, with 100\% denoting sitting and 0\% indicating standing.
Figure \ref{fig:pressure_vs_time} shows the pressure change in both PAMs simultaneously, since the pressure sensor is connected to both PAMs (see Fig.~\ref{fig:pneumatics}). 
The intervals when the pump was switched on and off are indicated, as well as the amount of recovered air.
In both figures, the exoskeleton states are indicated by numbers. The experiment is divided into seven parts as follows. 

\textbf{Initial Sitting:}
The experiment started by equalizing the actuator to atmospheric pressure in \circled{1}. The air pump gradually filled the PAM to a set pressure of \mbox{3.2 bar} in \circled{2}. In \circled{3} the air pump was switched off as the PAM was charged with compressed air but had not yet been released into the cylinder and therefore the exoskeleton was still transparent.

\textbf{Standing Up:} In \circled{4}, the sensors detected the subject's standing up movement, thus the actuator released the compressed air into the pneumatic cylinder, providing an active torque of \mbox{$\approx$6.7 Nm} which gradually decreased as the pressure dropped until \circled{5} (fully standing), at which point the pressure was equal to $\approx$1.85 bar and the torque $\approx$2 Nm. 

\textbf{Sitting Down:}
Upon sensor detection of the standing position, the mode was changed to "Quasi-passive damper". Thus, from \circled{5} to \circled{6}, when the subject transitioned to sitting, the air was compressed exclusively in the pneumatic cylinder, reaching a maximum torque of \mbox{$\approx$14.88 Nm}. 

\textbf{Air Return:}
By detecting the subjects' full seating in \circled{6}, and by positioning valve 2 to '1', compressed air from the cylinder had flown back to the PAM, restoring the PAM pressure from $\approx$1.85 bar to $\approx$2.5 bar, as the subject went from standing to sitting while the air pump was off. 

\textbf{Transparent Sitting:}
In \circled{7}, "Hold Air, Transparent" state holds the remaining air in the PAM but releases the compressed air from the cylinder allowing unobstructed sitting. 

\textbf{Next Repetition:}
In the next repetition, the air pump only had to compensate for air loss, as the joint  hadn't fully rotated to its end position. Thus, the air pump has been restarted (state \circled{2}), which allowed compensation for the pressure difference in the PAM up to \mbox{3.2 bar}. 

\textbf{End:}
Finally, all the compressed air from the actuator was released in \circled{1} and the experiment was completed.
\section{Discussion}

\begin{figure}[ht!]
        \centering 
    \includegraphics[clip,trim=0cm 0cm 0cm 0cm,width=0.45\textwidth]{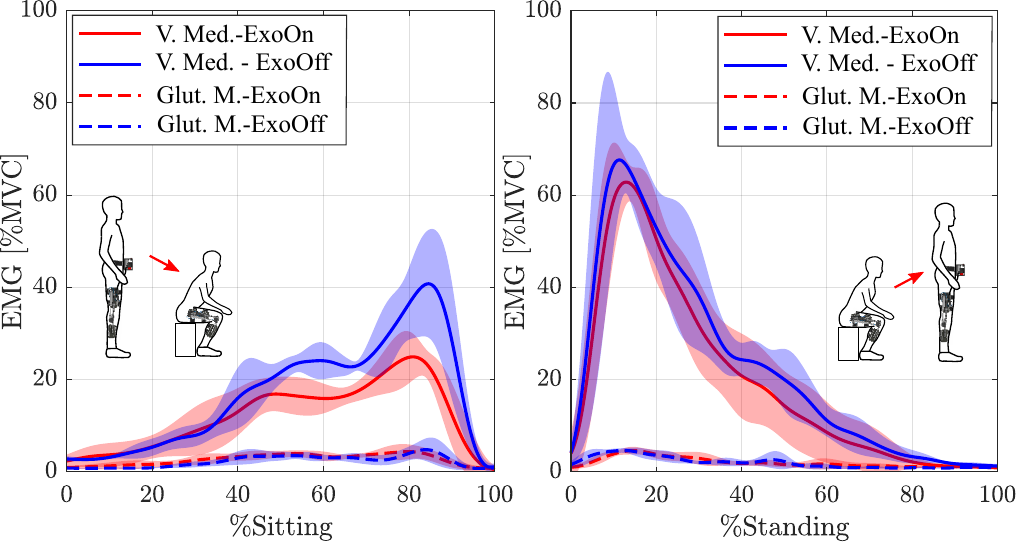}
    \caption{EMG recordings of the vastus medialis and gluteus maximus during sitting and standing transfers under assisted and unassisted conditions. 
    }\label{fig:results_1}
\end{figure}

This section discusses the energy recovery of the exoskeleton with the simultaneous muscle activity reduction in sitting transfers. 
Fig.~\ref{fig:results_1} displays average muscle activity over ten repetitions for transitions between sitting and standing, comparing unassisted (ExoOff) and assisted (ExoOn) conditions. The mean envelope of the raw EMG signals was extracted to analyze muscle effort, involving a series of filtering steps: $4^{\text{th}}$ order bandpass Butterworth (20-400 Hz), rectification, and low pass filtering (5 Hz, $4^{\text{th}}$ order Butterworth). EMG amplitude was normalized to the subject's MVC, segmented, and averaged based on \%Sitting and \%Standing identified from the knee angle.
\textbf{EMG reduction with simultaneous energy recovery:}
Interestingly, returning compressed air to the exoskeleton reservoir not only saved energy and pumping time but also reduced muscular effort. Notably, V. Med. activity decreased by $\approx$31\% (mean effort) when sitting down with assistance compared to unassisted. Glut. M. showed no major changes between the conditions, with low activity normalized to MVC. During standing up, a mean effort was reduced by $\approx$13\% in V. Med. when the exoskeleton assisted.

Actuation frequency relies on the air pump and the desired pressure, set at 3.2 bar during experimentation.
To assess energy recovery usefulness, three states with specific pressures are highlighted: state \circled{3} at 3.2 bar, state \circled{5} at $1.85\pm 0.01$ bar, and state \circled{7} at $2.46\pm 0.01$ bar. With energy recovery, the air pump has to refill only from $2.46\pm 0.01$ up to 3.2 bar. Without energy recovery, the air pump has to refill from the residual PAM pressure after actuation, $1.85\pm 0.01$, up to 3.2 bar.
The time to restore pressure to the desired level is determined using Eq. (\ref{eq:pm}), derived earlier from air pump characterization. The results are listed in the following table. 

\begin{table}[h!]
\caption{Max. Actuation frequency w/ and w/o Energy Recovery}
\begin{threeparttable}
\begin{tabular}{|P{2cm}|P{2.49cm}|P{2.49cm}|}
 \hline
\scriptsize{} & \scriptsize{Air pump: BD-04A-20L} & \scriptsize{{Air pump: BD-07A-35L}}\\ 
\hline
\scriptsize{Max. Freq. w/ ER\tnote{1}}& \scriptsize{$8.32\pm 0.04$ $\text{min}^{-1}$}& \scriptsize{$71.53\pm 0.66$ $\text{min}^{-1}$}\\
\hline
\scriptsize{Max. Freq. w/o ER\tnote{2}}& \scriptsize{$6.55\pm 0.02$ $\text{min}^{-1}$}& \scriptsize{$43.31\pm 0.34$ $\text{min}^{-1}$}\\

\hline
\end{tabular}
\begin{tablenotes}
\scriptsize{\item[1] Time to restore pressure with BD-04A-20L: $3.61\pm 0.02$ s (1 leg), $7.21\pm 0.03$s (both legs). With BD-07A-35L: $0.42\pm 0$ s (1 leg), $0.84\pm 0.01$ s (both)}
\scriptsize{\item[2] Time to restore pressure with BD-04A-20L: $4.58\pm 0.02$ s (1 leg), $9.16\pm 0.03$s (both legs). With BD-07A-35L: $0.71\pm 0.01$ s (1 leg), $1.42\pm 0.01$ s (both)}

\end{tablenotes}
\end{threeparttable}
\label{table_mechanism}
\end{table}

Yet, this frequency is derived from the air pump's ability. In experiments, the frequency, of course, is also influenced by the user's pace and transfer intervals. 
When pumping intervals are translated into energy savings through recovery, the exoskeleton's battery endurance extends by $1.27\pm 0.01$, $\approx$27\% (small pump), or $1.69\pm 0.02$, $\approx$69\% (large pump).

\section{Conclusion}

In conclusion, the key innovation of the JSI-KneExo is simultaneous human assistance with energy recovery without compromising transparency. Other innovations are pneumatic circuit design that provides different operation modes, portable pneumatic actuation using PAM as an air tank, and low energy consumption by operating either in quasi-passive or active mode. 
Future endeavors will delve into investigating the efficacy of JSI-KneExo 
in assisting with walking and analyzing the effects on a larger sample of subjects.

\ifCLASSOPTIONcaptionsoff
  \newpage
\fi

\bibliographystyle{IEEEtran}
\bibliography{references}

\begin{thebibliography}{10}
\providecommand{\url}[1]{#1}
\csname url@samestyle\endcsname
\providecommand{\newblock}{\relax}
\providecommand{\bibinfo}[2]{#2}
\providecommand{\BIBentrySTDinterwordspacing}{\spaceskip=0pt\relax}
\providecommand{\BIBentryALTinterwordstretchfactor}{4}
\providecommand{\BIBentryALTinterwordspacing}{\spaceskip=\fontdimen2\font plus
\BIBentryALTinterwordstretchfactor\fontdimen3\font minus
  \fontdimen4\font\relax}
\providecommand{\BIBforeignlanguage}[2]{{%
\expandafter\ifx\csname l@#1\endcsname\relax
\typeout{** WARNING: IEEEtran.bst: No hyphenation pattern has been}%
\typeout{** loaded for the language `#1'. Using the pattern for}%
\typeout{** the default language instead.}%
\else
\language=\csname l@#1\endcsname
\fi
#2}}
\providecommand{\BIBdecl}{\relax}
\BIBdecl

\bibitem{DonaldA.Neumann2016KinesiologyRehabilitation}
{Donald A. Neumann}, \emph{{"Kinesiology of the Musculoskeletal System
  Foundations for Rehabilitation"}}.\hskip 1em plus 0.5em minus 0.4em\relax
  Elsevier, 2016.

\bibitem{Donelan2008BiomechanicalEffort}
J.~M. Donelan, Q.~Li, V.~Naing, J.~A. Hoffer, D.~J. Weber, and A.~D. Kuo,
  ``{Biomechanical energy harvesting: Generating electricity during walking
  with minimal user effort},'' \emph{Science}, vol. 319, no. 5864, pp.
  807--810, 2008.

\bibitem{Wu2022GeneratingExoskeleton}
X.~Wu, W.~Cao, H.~Yu, Z.~Zhang, Y.~Leng, and M.~Zhang, ``{Generating
  Electricity During Locomotion Modes Dominated by Negative Work via a Knee
  Energy-Harvesting Exoskeleton},'' pp. 1--11, 2022.

\bibitem{Laschowski2019Lower-LimbReview}
B.~Laschowski, J.~McPhee, and J.~Andrysek, ``{Lower-Limb Prostheses and
  Exoskeletons With Energy Regeneration: Mechatronic Design and Optimization
  Review},'' \emph{Journal of Mechanisms and Robotics}, vol.~11, no.~4, pp.
  1--43, 2019.

\bibitem{Dollar2008DesignRunning}
A.~M. Dollar and H.~Herr, ``{Design of a quasi-passive knee exoskeleton to
  assist running},'' \emph{2008 IEEE/RSJ International Conference on
  Intelligent Robots and Systems, IROS}, pp. 747--754, 2008.

\bibitem{Collins2015ReducingExoskeleton}
S.~H. Collins, M.~Bruce~Wiggin, and G.~S. Sawicki, ``{Reducing the energy cost
  of human walking using an unpowered exoskeleton},'' \emph{Nature}, vol. 522,
  no. 7555, pp. 212--215, 2015.

\bibitem{DiNatali2020PneumaticExoskeleton}
C.~Di~Natali, A.~Sadeghi, A.~Mondini, E.~Bottenberg, B.~Hartigan, A.~De~Eyto,
  L.~O'Sullivan, E.~Rocon, K.~Stadler, B.~Mazzolai, D.~G. Caldwell, and
  J.~Ortiz, ``{Pneumatic Quasi-Passive Actuation for Soft Assistive Lower Limbs
  Exoskeleton},'' \emph{Frontiers in Neurorobotics}, vol.~14, no. June, pp.
  1--18, 2020.

\bibitem{Miskovic2022PneumaticApplications}
L.~Miskovic, M.~Dezman, and T.~Petric, ``{Pneumatic Quasi-Passive Variable
  Stiffness Mechanism for Energy Storage Applications},'' \emph{IEEE Robotics
  and Automation Letters}, vol.~7, no.~2, pp. 1705--1712, 2022.

\bibitem{Kumar2020ExtremumExoskeleton}
S.~Kumar, M.~R. Zwall, E.~A. Bolivar-Nieto, R.~D. Gregg, and N.~Gans,
  ``{Extremum Seeking Control for Stiffness Auto-Tuning of a Quasi-Passive
  Ankle Exoskeleton},'' \emph{IEEE Robotics and Automation Letters}, vol.~5,
  no.~3, pp. 4604--4611, 2020.

\bibitem{Sutrisno2019EnhancingExoskeletons}
A.~Sutrisno and D.~J. Braun, ``{Enhancing Mobility with Quasi-Passive Variable
  Stiffness Exoskeletons},'' \emph{IEEE Transactions on Neural Systems and
  Rehabilitation Engineering}, vol.~27, no.~3, pp. 487--496, 2019.

\bibitem{Walsh2007AAugmentation}
C.~J. Walsh, K.~Endo, and H.~Herr, ``{A quasi-passive leg exoskeleton for
  load-carrying augmentation},'' \emph{International Journal of Humanoid
  Robotics}, vol.~4, no.~3, pp. 487--506, 2007.

\bibitem{Laschowski2021SimulationRegeneration}
B.~Laschowski, R.~S. Razavian, and J.~McPhee, ``{Simulation of Stand-to-Sit
  Biomechanics for Robotic Exoskeletons and Prostheses with Energy
  Regeneration},'' \emph{IEEE Transactions on Medical Robotics and Bionics},
  vol.~3, no.~2, pp. 455--462, 2021.

\bibitem{Schmidt2017TheTransfers}
K.~Schmidt, J.~E. Duarte, M.~Grimmer, A.~Sancho-Puchades, H.~Wei, C.~S.
  Easthope, and R.~Riener, ``{The myosuit: Bi-articular anti-gravity exosuit
  that reduces hip extensor activity in sitting transfers},'' \emph{Frontiers
  in Neurorobotics}, vol.~11, no. OCT, pp. 1--16, 2017.

\bibitem{Shepherd2017DesignAssistance}
M.~K. Shepherd and E.~J. Rouse, ``{Design and Validation of a
  Torque-Controllable Knee Exoskeleton for Sit-to-Stand Assistance},''
  \emph{IEEE/ASME Transactions on Mechatronics}, vol.~22, no.~4, pp.
  1695--1704, 2017.

\bibitem{Liu2021LowSit-to-Stand}
L.~Liu, Z.~Hong, B.~Penzlin, B.~J. Misgeld, C.~Ngo, L.~Bergmann, and
  S.~Leonhardt, ``{Low Impedance-Guaranteed Gain-Scheduled GESO for
  Torque-Controlled VSA with Application of Exoskeleton-Assisted
  Sit-to-Stand},'' \emph{IEEE/ASME Transactions on Mechatronics}, vol.~26,
  no.~4, pp. 2080--2091, 2021.

\bibitem{Zhu2021DesignOrthoses}
H.~Zhu, C.~Nesler, N.~Divekar, V.~Peddinti, and R.~D. Gregg, ``{Design
  Principles for Compact, Backdrivable Actuation in Partial-Assist Powered Knee
  Orthoses},'' \emph{IEEE/ASME Transactions on Mechatronics}, vol.~26, no.~6,
  pp. 3104--3115, 2021.

\bibitem{Vantilt2019Model-basedMovements}
J.~Vantilt, K.~Tanghe, M.~Afschrift, A.~K. Bruijnes, K.~Junius, J.~Geeroms,
  E.~Aertbeli{\"{e}}n, F.~De~Groote, D.~Lefeber, I.~Jonkers, and
  J.~De~Schutter, ``{Model-based control for exoskeletons with series elastic
  actuators evaluated on sit-to-stand movements},'' \emph{Journal of
  NeuroEngineering and Rehabilitation}, vol.~16, no.~1, pp. 1--21, 2019.

\bibitem{Jamsek2020GaussianExoskeletons}
M.~Jam{\v{s}}ek, T.~Petri{\v{c}}, and J.~Babi{\v{c}}, ``{Gaussian mixture
  models for control of Quasi-passive spinal exoskeletons},'' \emph{Sensors
  (Switzerland)}, vol.~20, no.~9, pp. 1--13, 2020.

\bibitem{Maeda2012MuscleMuscles}
D.~Maeda, K.~Tominaga, T.~Oku, H.~T. Pham, S.~Saeki, M.~Uemura, H.~Hirai, and
  F.~Miyazaki, ``{Muscle synergy analysis of human adaptation to a
  variable-stiffness exoskeleton: Human walk with a knee exoskeleton with
  pneumatic artificial muscles},'' \emph{IEEE-RAS International Conference on
  Humanoid Robots}, no.~1, pp. 638--644, 2012.

\bibitem{Knaepen2014Human-robotExoskeleton}
K.~Knaepen, P.~Beyl, S.~Duerinck, F.~Hagman, D.~Lefeber, and R.~Meeusen,
  ``{Human-robot interaction: Kinematics and muscle activity inside a powered
  compliant knee exoskeleton},'' \emph{IEEE Transactions on Neural Systems and
  Rehabilitation Engineering}, vol.~22, no.~6, pp. 1128--1137, 2014.

\bibitem{Miskovic2023PneumaticEvaluation}
L.~Miskovic, M.~Dezman, and T.~Petric, ``{Pneumatic Exoskeleton Joint with a
  Self-Supporting Air Tank and Stiffness Modulation: Design, Modeling, and
  Experimental Evaluation},'' \emph{TechRxiv. Preprint.}, 2023.

\bibitem{Wehner2013AAssistanceb}
M.~Wehner, B.~Quinlivan, P.~M. Aubin, E.~Martinez-Villalpando, M.~Baumann,
  L.~Stirling, K.~Holt, R.~Wood, and C.~Walsh, ``{A lightweight soft exosuit
  for gait assistance},'' \emph{Proceedings - IEEE International Conference on
  Robotics and Automation}, pp. 3362--3369, 2013.

\bibitem{Heo2020BackdrivableAssistance}
U.~Heo, S.~J. Kim, and J.~Kim, ``{Backdrivable and Fully-Portable Pneumatic
  Back Support Exoskeleton for Lifting Assistance},'' \emph{IEEE Robotics and
  Automation Letters}, vol.~5, no.~2, pp. 2047--2053, 2020.

\bibitem{Martens2017ModelingModels}
M.~Martens and I.~Boblan, ``{Modeling the static force of a Festo pneumatic
  muscle actuator: A new approach and a comparison to existing models},''
  \emph{Actuators}, vol.~6, no.~4, pp. 1--11, 2017.

\bibitem{Chou1996MeasurementMuscles}
C.~P. Chou and B.~Hannaford, ``{Measurement and modeling of McKibben pneumatic
  artificial muscles},'' \emph{IEEE Transactions on Robotics and Automation},
  vol.~12, no.~1, pp. 90--102, 1996.

\bibitem{Hermens2000DevelopmentProcedures}
H.~J. Hermens, B.~Freriks, C.~Disselhorst-Klug, and G.~Rau, ``{Development of
  recommendations for SEMG sensors and sensor placement procedures},'' \emph{J.
  Electromyogr. Kinesiol}, vol.~10, no.~5, p. 361–374, 2000.

\end{thebibliography}

\end{document}